\documentclass{article}

    \usepackage[final]{neurips_2019}

\usepackage[utf8]{inputenc} 
\usepackage[T1]{fontenc}    
\usepackage{url}            
\usepackage{booktabs}       
\usepackage{amsfonts}       
\usepackage{nicefrac}       
\usepackage{microtype}      

\usepackage{natbib}
\usepackage{graphics} 
\usepackage{epsfig} 
\usepackage{amsmath} 
\usepackage{amssymb}  
\usepackage{graphicx}
\usepackage{caption}
\usepackage{float}
\usepackage{subfigure}
\usepackage{xcolor}
\usepackage{hyperref}
\hypersetup{
    colorlinks =false,
    urlcolor=false,
    citebordercolor=white,
    linkbordercolor = white,
}

\title{The iCub multisensor datasets for robot and computer vision applications}

\author{  Murat Kirtay\\
  Adaptive Systems Group, Department of Computer Science\\
  Humboldt-Universität zu Berlin\\
  Berlin, Germany\\
  \texttt{murat.kirtay@informatik.hu-berlin.de} 
   \And
  Ugo Albanese, Lorenzo Vannucci, Guido Schillaci,
 Cecilia Laschi and Egidio Falotico \\
  The BioRobotics Institute, Scuola Superiore Sant'Anna \\
  Pontedera (PI), Italy \\
  \texttt{\{u.albanese,l.vannuci,g.schillaci,c.lashi,e.falotico\}@santannapisa.it} }

\begin{document}

\maketitle

\begin{abstract}

This document presents novel datasets, constructed by employing the iCub robot equipped with an additional depth sensor and color camera. We used the robot to acquire color and depth information for 210  objects in different acquisition scenarios. At this end, the results were large scale datasets for robot and computer vision applications: object representation, object recognition and classification, and action recognition.

\end{abstract}

\section{Introduction}
Multisensory integration is an essential skill that biological agents (e.g., humans) can perform with apparent ease to achieve a considerable number of vision-based tasks~\citep{lahat2015multimodal, Ramachandram2017survey}. Among these tasks, there are: recognizing an object within a cluttered scene, detecting and tracking an object form various poses and distances, recognizing the action of others while working on the same task. Recent advancements in machine learning and robotics, thanks to the availability of computational power and inexpensive sensors, provided significant contributions to state-of-the-art results in the domain of machine vision. Yet, learning how to combine multiple sensory information to perform given tasks is still a challenging research problem for artificial agents (e.g., robots, or decision support systems). 

Processing a large scale multisensor dataset is one of the crucial components to achieve a vision-based task in which discovering complementary characteristics of the employed sensors are critical. Therefore, in this report, we present novel (and large scale) multisensor datasets which are suitable for robot and computer vision applications. Although we constructed these datasets for vision-related tasks such as object recognition, representation, and action recognition, the datasets can also be employed into various cognitive robotics applications: multimodal object concept formation, multimodal object learning, and affordance learning.

Unlike existing datasets, the datasets reported in this paper present the following advantages to the interested researchers. First, some of these datasets were collected in a single (or similar) experimental setting to address particular problems such as object recognition and categorization~\citep{Araki2011concept, Fanello2013iCubWF}. Second, the datasets provided in~\citep{lomonaco2017Core, lai2011large} constructed by using depth and color cameras without employing a robotic agent. In that, the machine learning models trained in these datasets might not be extended to the robot vision problems.  Lastly, our datasets can be considered as a large scale regarding the number of objects and employed sensory data. Additionally, we provide our datasets in an open-access format to enable the researchers to benchmark their models and reproduce our base-line results. Although we, here, compared our datasets with the representative multimodal datasets in the literature, the more comprehensive report can be found in~\citep{ZHANG201686}.

In this report, we aim at presenting the dataset acquisition setups, the dataset constructing pipelines for each scenario, and the possible robot and computer vision problems for which each of these datasets can be useful.

\section{Data acquisition setup}
\begin{figure}[ht]
	\centering
	\begin{center}
	
	  \subfigure[The iCub robot ]{\label{expsetup:a}\includegraphics[width=0.35\textwidth, height=0.196\textheight]{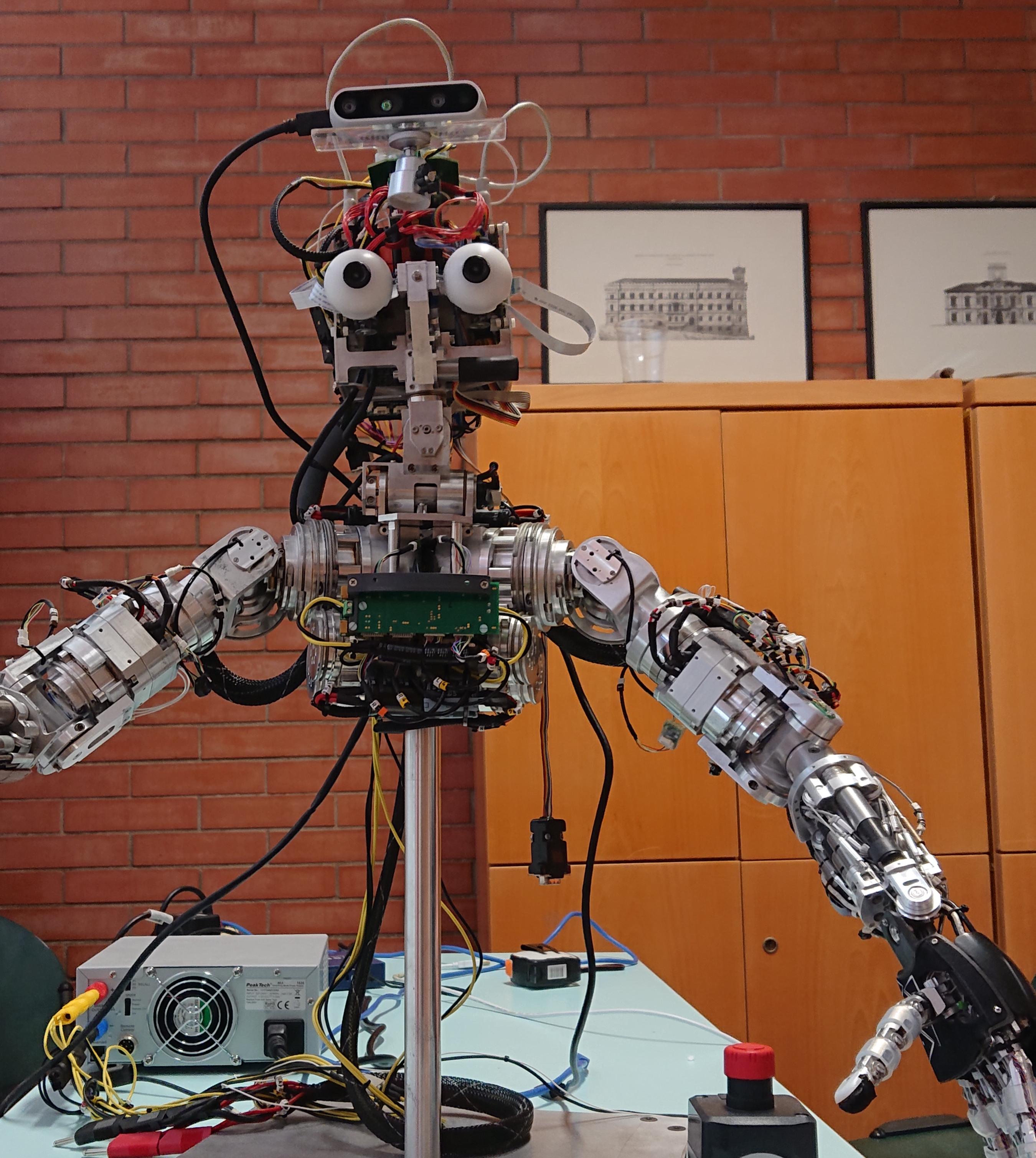}}
	  \subfigure[Objects ]{\label{expsetup:b}\includegraphics[width=0.5\textwidth]{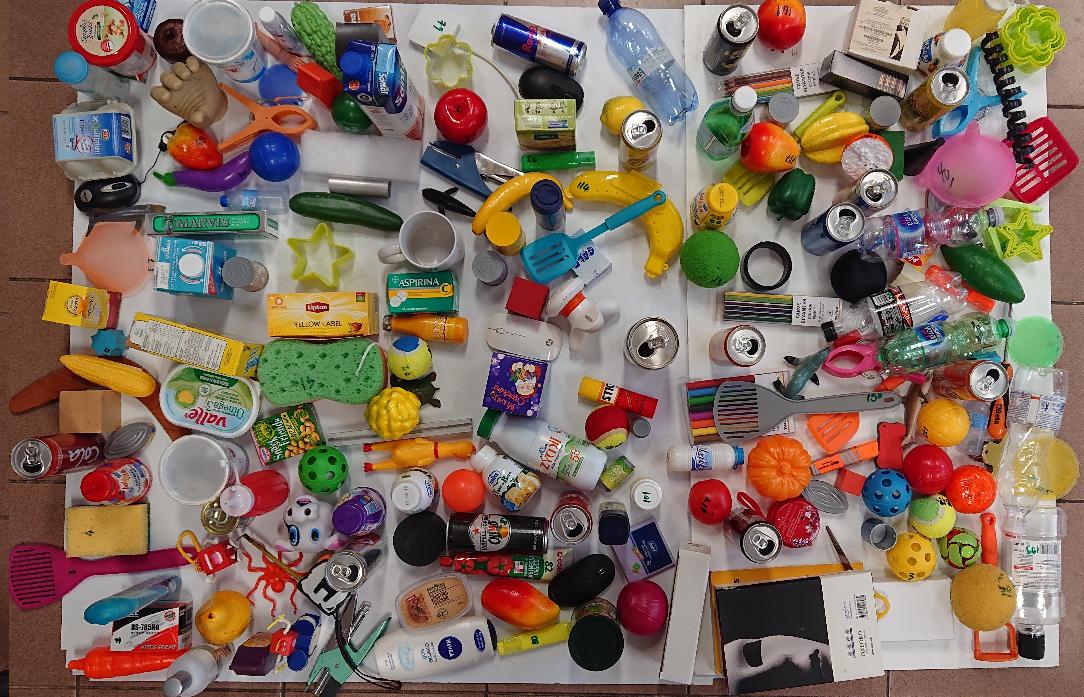}}	

	\end{center}
	\caption{Experiment setup: the iCub robot and objects}
	\label{fig:expsetup}
\end{figure}

In this section, we describe the acquisition setup and provided a brief overview of each acquisition scenario. To construct the datasets, we employed the iCub robot -- equipped with three RGB cameras and a depth sensor-- in four different scenarios, with 210 daily life objects, including toys, kitchen equipment, boxes, etc. Figure~\ref{expsetup:a} shows the iCub robot, and Figure~\ref{expsetup:b} presents the objects used for dataset construction. To be concrete, we designed a data acquisition pipeline to obtain color and depth information via  Dragonfly cameras mounted on the robot’s eyes, and an Intel Realsense d435i camera located above the eyes. The pipeline was developed by integrating RealSense (i.e., librealsense) libraries~\footnote{\href{https://github.com/IntelRealSense/librealsense}{https://github.com/IntelRealSense/librealsense}} and the iCub’s middleware (YARP)~\citep{metta2006yarp}. In this setup, we captured color and depth images with a size of $640\times480$.  The color images contain the RGB matrices, and the depth images contain the normalized distance between each pixel in the scene and the place where the depth sensor is located. Note that, in this report, we visualized the depth data as images by using color maps to distinguish different distance in the images.

Table~\ref{table:dataset} presents the general information about each scenario in terms of a number of objects used in the experiment and captured images via sensors. In this table, the data acquisition scenario placed in the first column and the corresponding number of the objects used in the experiment is provided in the second column. The number of images collected per each sensor and the total number of images in the corresponding dataset are presented in the third and last columns of the Table~\ref{table:dataset}. In these columns, IR depth and IR Color refer to the depth images obtained via Intel RealSense sensor. Similarly, iCub left and iCub right indicate the color cameras located in the robot’s left and right eye, respectively.
\begin{table}[ht!]
\centering
\begin{tabular}{lcccccc}
\toprule
Dataset & Objects & \multicolumn{4}{c}{Number of images per sensor} & Total 	
  \\
\midrule
{}       & {}        & IR Depth       & IR Color   & iCub Left   & iCub Right   & {}\\ 
RGB-D turntable  & $210$  & $15120$ & $15120$ & $15120$ & $15120$ & $60480$    \\ 
RGB-D operator   & $210$  & $189000$  & $189000$   & $189000$   & $189000$  & $756000$    \\ 
Scene understanding & $210$  & $860$  & $860$   & $860$   & $860$  & $3440$    \\ 
Action recognition & $20$  & $6400$  & $6400$   & $6400$   & $6400$  & $25600$    \\ 

\bottomrule
\end{tabular}
\\

\caption{Dataset specification for each sensors where IR refers to Intel Realsense d435i.}
\label{table:dataset}
\end{table}

Based on the number of images in the dataset, we can affirm that the size of the whole experiment data can be considered as large scale, while the inclusion of both color and depth sensors makes this dataset multimodal. Due to these features, these datasets can be utilized for benchmarking purposes for machine learning algorithms, which can be integrated on the iCub robot. Related to this, one of our ongoing works is to provide benchmarking results for state-of-the-art recognition, classification, scene understanding, and detection methods: deep convolutional network, capsule networks, faster R-CNN, etc.

\section{An RGB-D dataset constructed with a motorized turntable (RGB-D turntable)} \label{turntable}
\begin{figure}[ht!]
	\centering
	\begin{center}
		\includegraphics[width=1\textwidth]{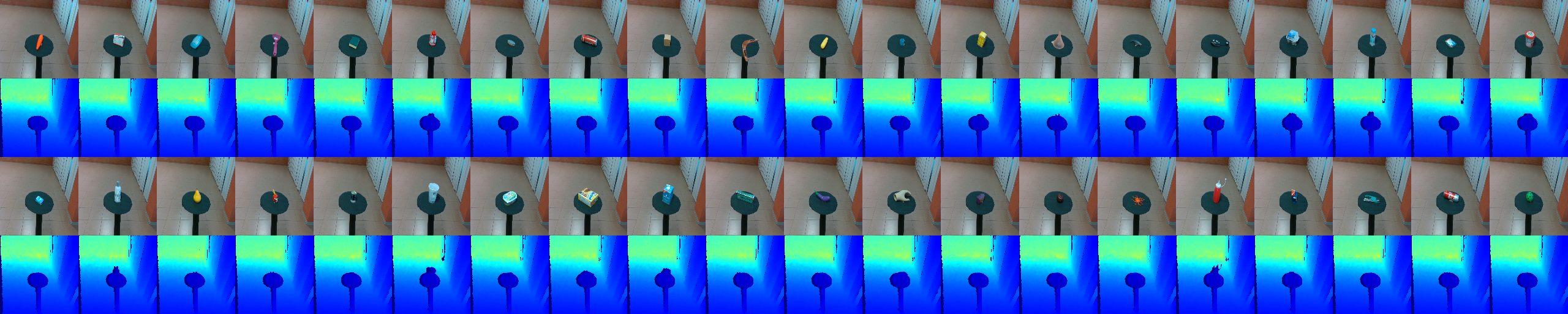}
	\end{center}
	\caption{Sample of depth and color images from the RGB-D turntable dataset.}
	\label{fig:turntable}
\end{figure}

In this scenario, the experiment setup consists of a motorized turntable and the iCub robot. In this setup, we designed an automatized acquisition pipeline  to capture images for the object in a similar scenario which is described in~\citep{nene1996coil100,kirtay2017dataset}. To capture color and depth images, we respectively put the objects on the turntable, and then the table was rotated by five degrees till it completed a full rotation. At the end of this experiment, we obtained 72 different views of an object and for each view we collected 72 depth and 216 color images. A sample of color and depth images of various objects is presented in Figure~\ref{fig:turntable}. In total, this dataset provides 60480 color and depth images that can be employed for computer and robot vision applications.

This dataset can be used in various computer vision tasks, including object representation, classification, and recognition. Moreover, since the depth and color images are mapped to the corresponding positions of the motorized turntable, this dataset might be employed to learning sensorimotor schemes in the context of robot and computer vision application.

\section{An RGB-D dataset constructed with human operators (RGB-D operator)}
\begin{figure}[ht!]
	\centering
	\begin{center}
		\includegraphics[width=1\textwidth]{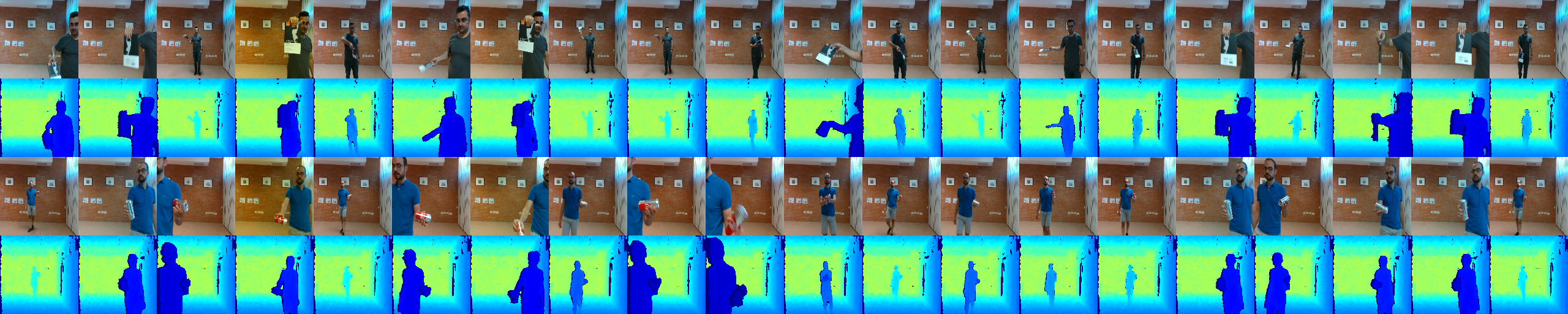}
	\end{center}
	\caption{Sample of depth and color images from various objects in the RGB-D operator dataset.}
	\label{fig:operator}
\end{figure}

In this acquisition scenario, we aimed at collecting different variations of an object, including distance to the robot, pose, illumination, etc. We conduct this experiment with the same setup that we introduced in Section~\ref{turntable}, with the exception of the turntable, replaced by an operator. By employing an operator, instead of a turntable, we obtained more different views of the object in three rotational axes, instead of a single one, with more poses, and under noisier environmental conditions such as specularities, shadows, etc. In this experiment, the operator holds an object and moves inside the field of view of the robot’s cameras. While doing so, the operator changes the pose of the object in random order, without being instructed. We noted that some of the images may not contain or partially contain the objects due to the random movement of the operator that may, partially or completely, occlude the object.  Figure~\ref{fig:operator} shows a sample of color and depth images for several objects from the robot's viewpoint. For a single object, we obtained 900 different views, including both color and depth sensors. In total, for each object, we captured 900 depth and 2700 color images. Overall, this dataset consists of 756000 images for 210 objects.

\newpage
\section{Scene understanding dataset}
\begin{figure}[ht!]
    \centering
    \begin{center}
        \includegraphics[width=1\textwidth]{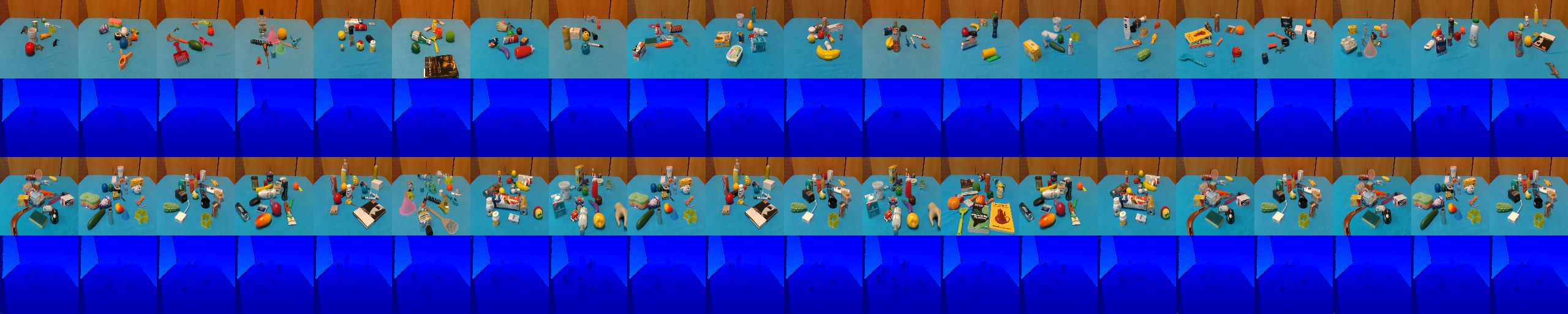}
    \end{center}
    \caption{Color and depth images of cluttered scenes with 10 and 20 objects, from the scene understanding dataset.}
    \label{fig:scene}
\end{figure}

In this experiment, we constructed two datasets to perform indoor scene understanding by creating a cluttered scene with 10 and 20 objects. In the first setup, we put 10 objects in front of the robot to capture depth and color images. After obtaining color and depth data of the scene, we then randomly shuffle objects’ poses 20 times to capture more variations of the scene. This procedure was repeated for all objects in the dataset. To this end, we created 21 different scenes with 10 objects and we collected 1680 depth and color images. In the second setup, we replicate the same procedure with 20 objects and we shuffled these objects in the scene 40 times, thus creating a dataset with 1760 color and depth images. In total, we collected 3340 color and depth images by combining the data from the first and second experiments. The first row of Figure~\ref{fig:scene} shows cluttered scenes with 10 objects, while the second row presents scenes constructed with 20 objects.
We envision that these datasets can be employed for an indoor scene understanding task to extract functional and semantic information about the scene. For instance, the iCub robot can detect common kitchen tools and generate text-to-speech description for the operator who can then perform a specific task (e.g., flipping a hamburger with a spatula) by requesting the appropriate tools from the robot.

\section{Action recognition dataset}
\begin{figure}[ht!]
    \centering
    \begin{center}
        \subfigure[Experiment setup ]{\label{toolsAndObjs:a}\includegraphics[width=0.42\textwidth, height=0.2\textheight]{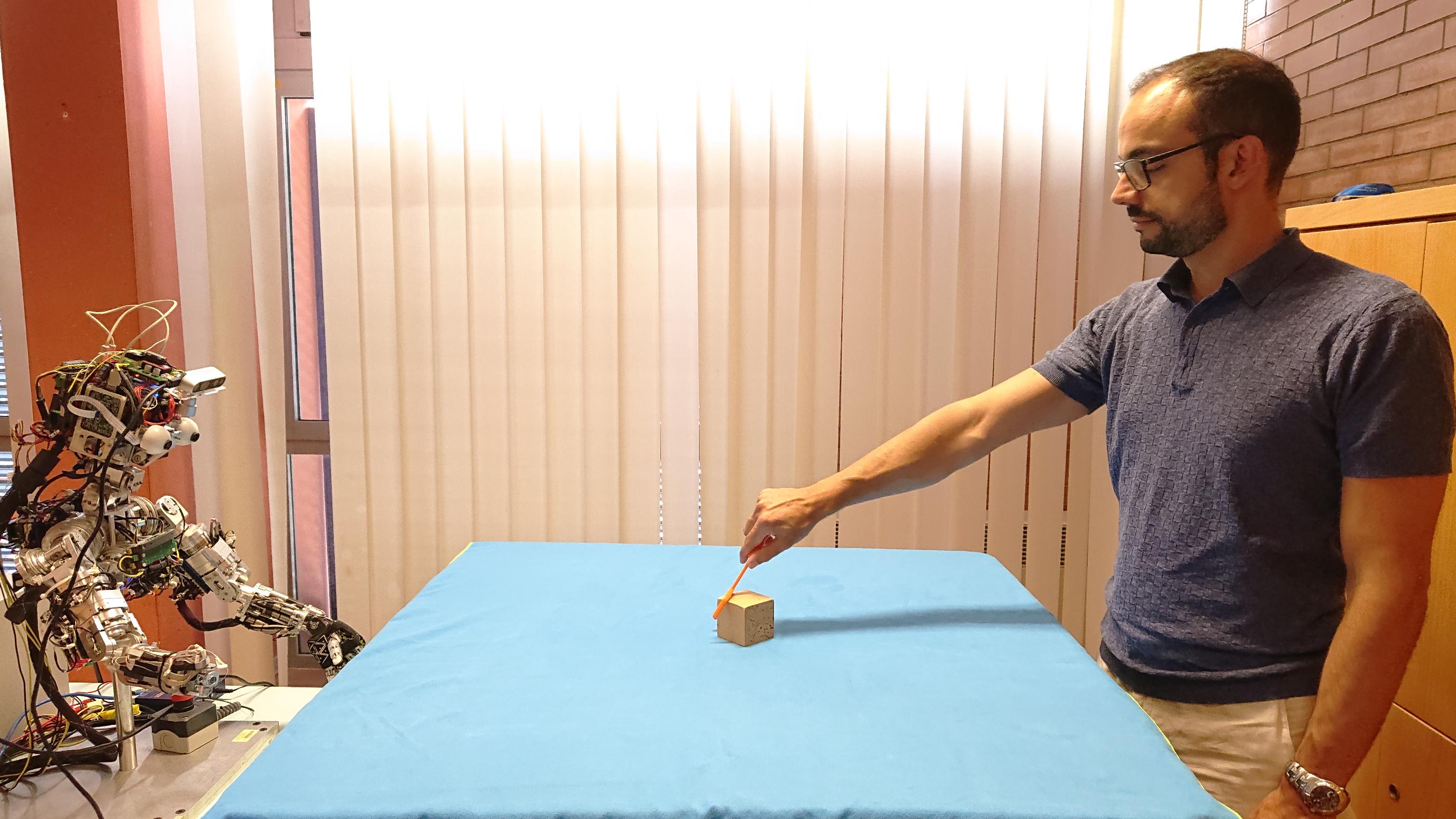}} 
        \subfigure[Objects ]{\label{toolsAndObjs:b}\includegraphics[width=0.25\textwidth, height=0.2\textheight]{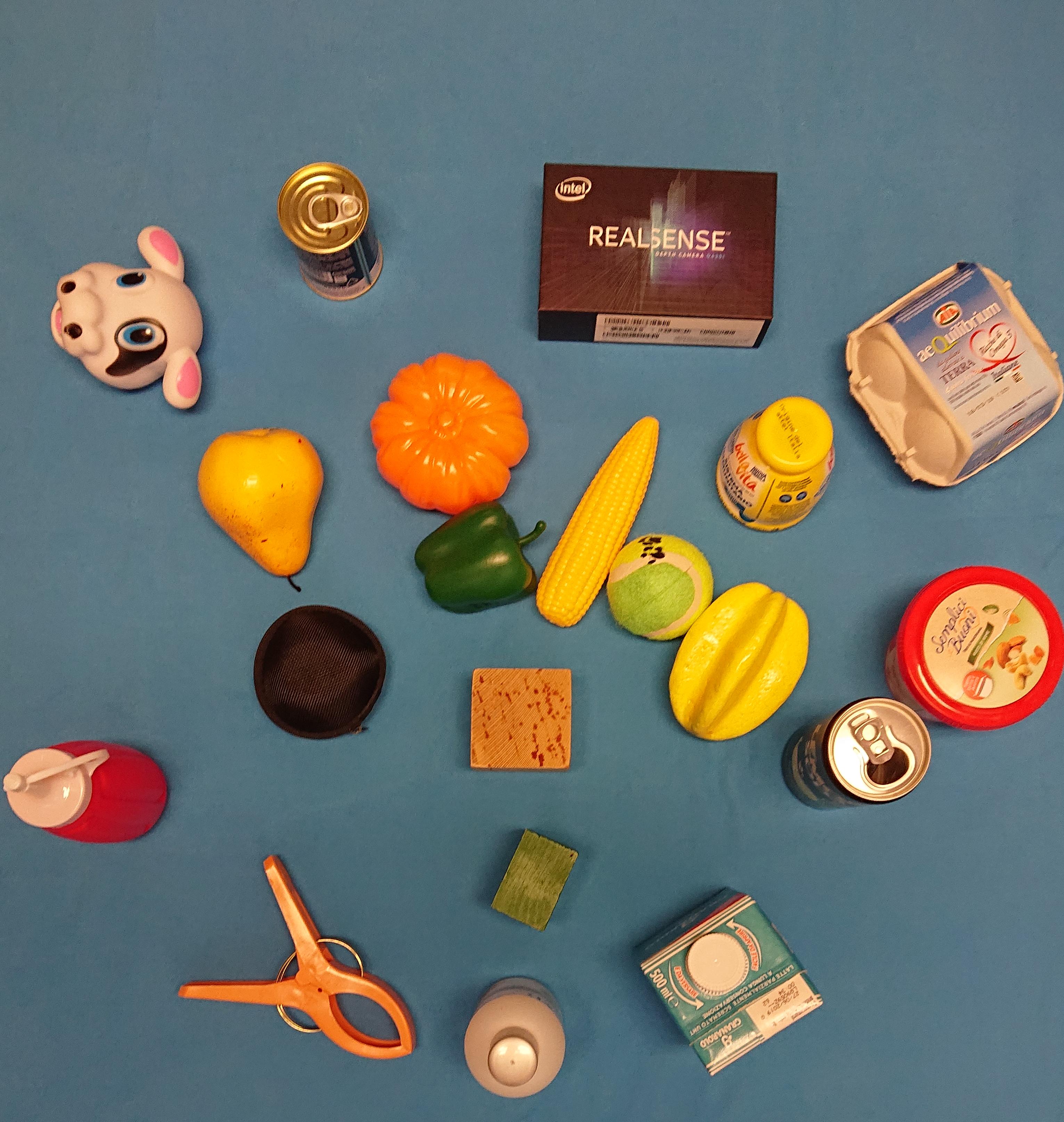}} 
        \subfigure[Tools ]{\label{toolsAndObjs:c}\includegraphics[width=0.25\textwidth, height=0.2\textheight]{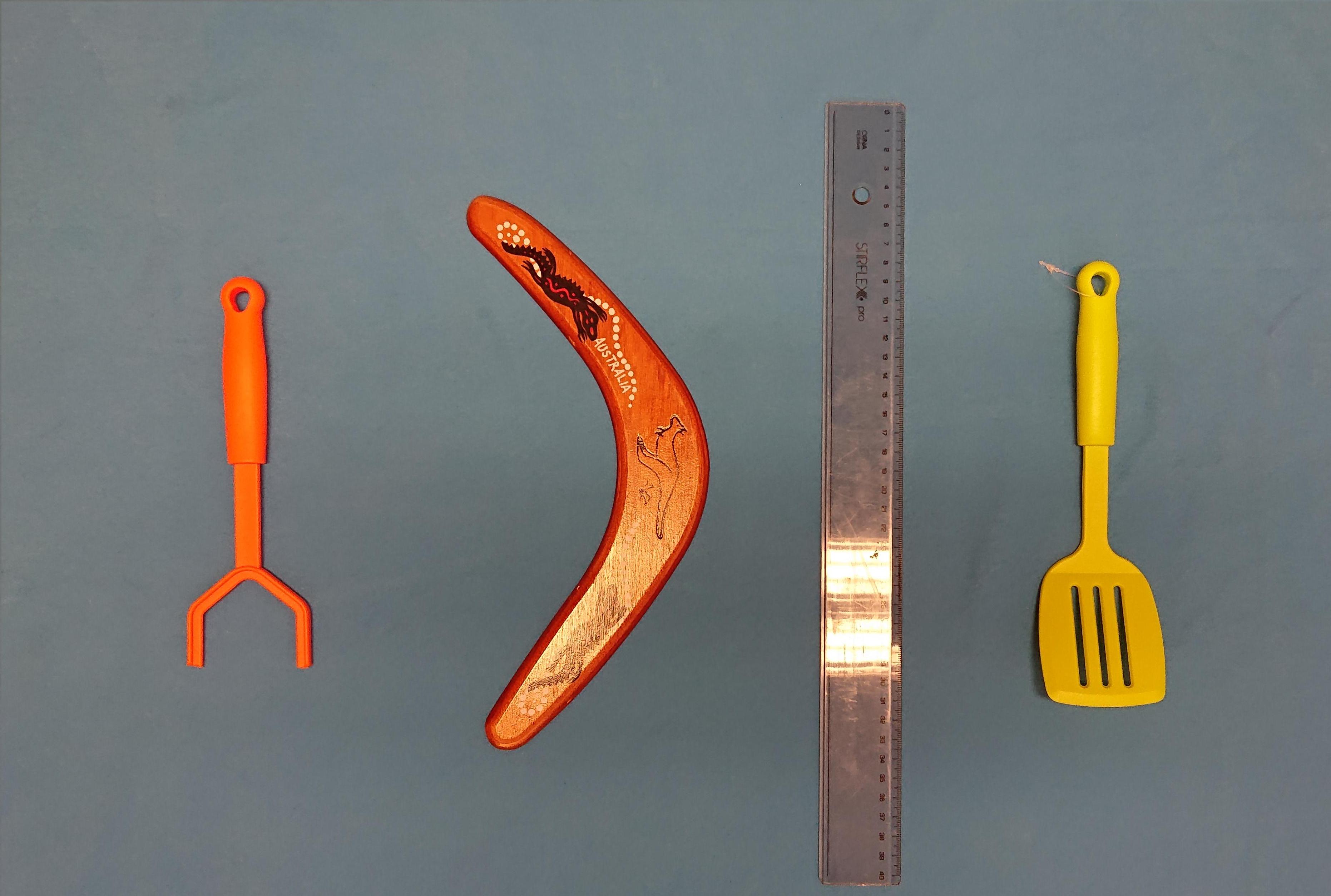}}

    \end{center}
    \caption{Experimental setup, objects and tools that we employed for the action recognition dataset}
    \label{fig:toolsAndObjs}
\end{figure}

In this scenario, we collect color and depth images related to human actions performed with specific tools on a set of objects. Concretely, we designed a setup where the iCub robot becomes an action-observer and four different operators have the role of be action-performers. Actions are performed on 20 objects --with different properties such as material, texture, size, etc.-- chosen among then 210 objects that we used for the previous datasets. As shown in Figure~\ref{toolsAndObjs:a}, the operator performs an action on the object by using one of the four available tools: hook, ruler, slingshot, and spatula. The objects and tools used in this setup are shown in Figure~\ref{toolsAndObjs:b},~\ref{toolsAndObjs:c}, respectively.

\begin{figure}[ht!]
    \centering
    \begin{center}
        \subfigure[Push  ]{\label{actions:a}\includegraphics[width=0.24\textwidth]{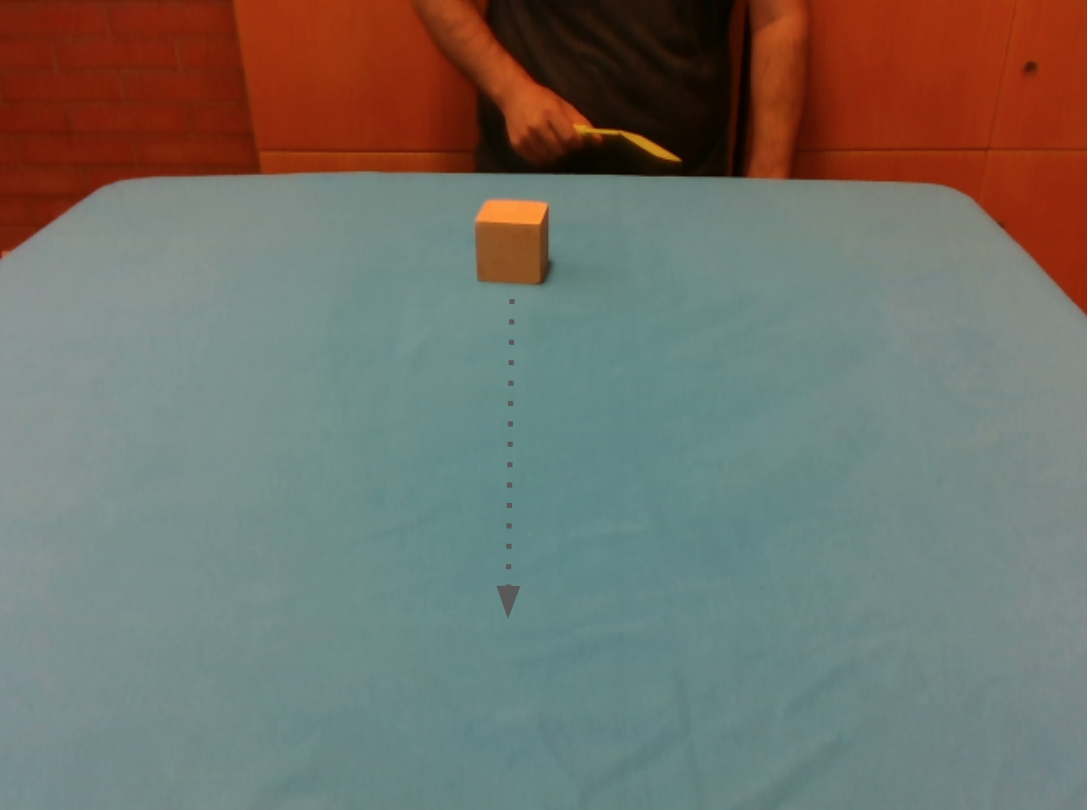}}
        \subfigure[Pull ]{\label{actions:b}\includegraphics[width=0.24\textwidth]{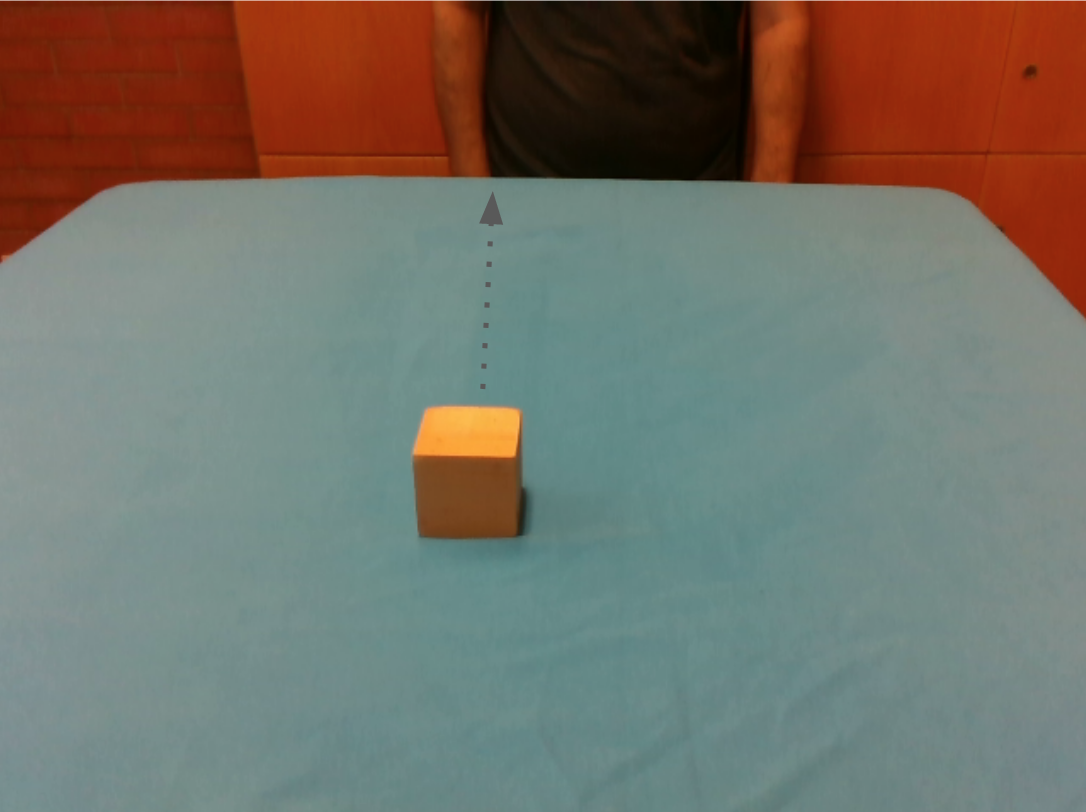}}
        \subfigure[Left to right]{\label{actions:c}\includegraphics[width=0.24\textwidth]{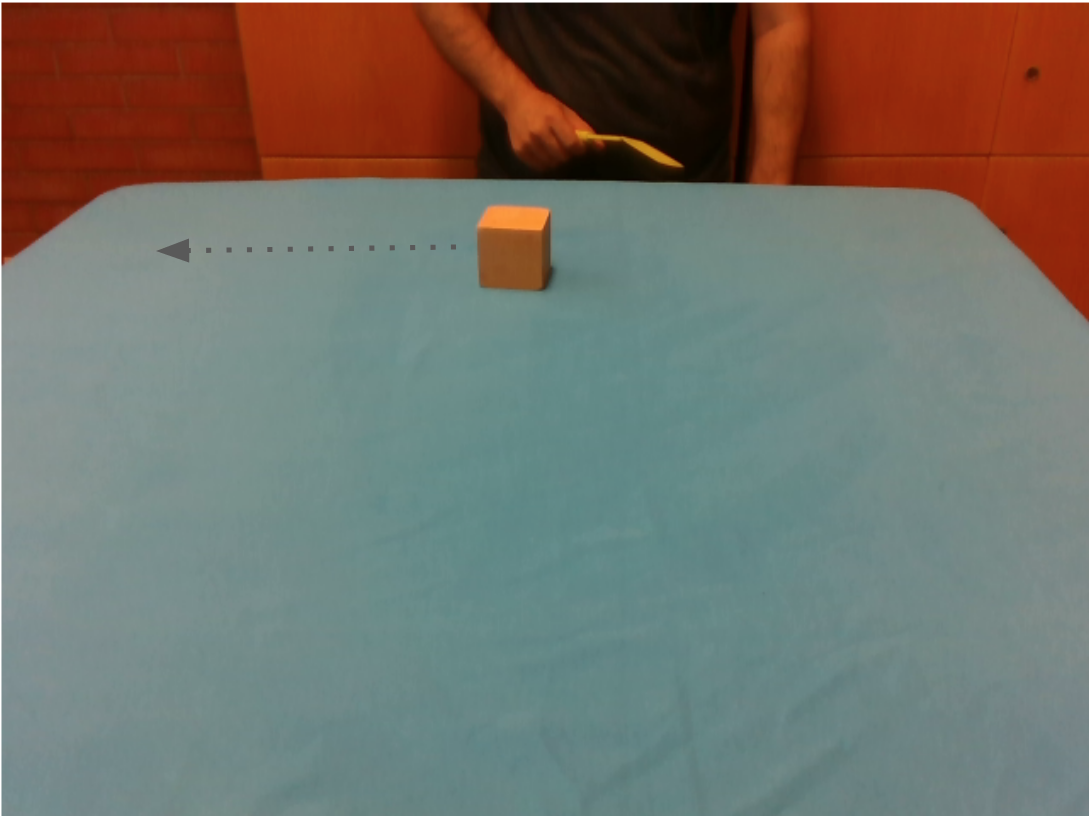}}
        \subfigure[Right to left]{\label{actions:d}\includegraphics[width=0.24\textwidth]{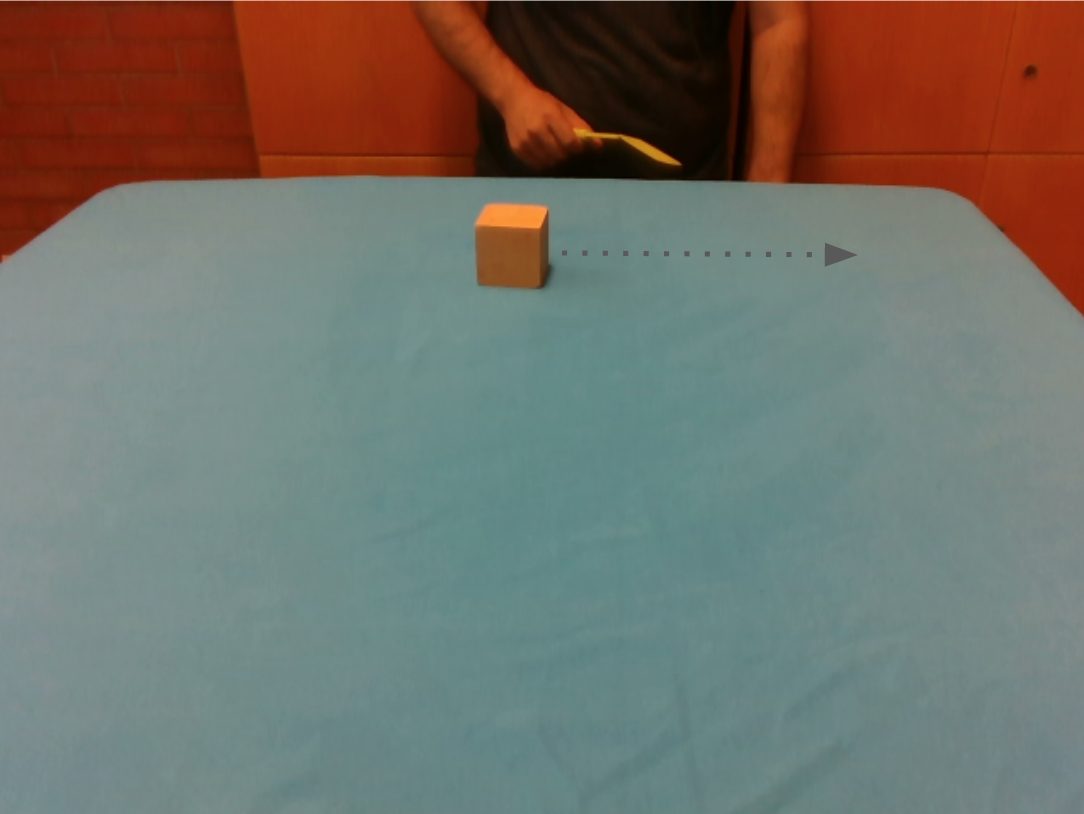}}
    
    \end{center}
    \caption{Illustration of actions performed by the operators to obtain the action recognition dataset.}
    \label{fig:actions}
\end{figure}

The operator uses every tool to perform four distinct actions: pulling, pushing, moving an object from left to right, and moving an object from right to left. In this setting, we captured depth and color images before and after performing every action. In doing so, we obtained the initial pose of the object on the scene, and the effect of the action applied on the object by the operator with the tool. The action and a possible movement paths of the object are illustrated in Figure~\ref{fig:actions}.

We describe the data acquisition procedure in the following way. First, the operator puts an object on a table in front of the robot. Then, the depth and color images are captured for the initial pose of the object. Next, the operator chooses a tool and performs an action on the object. After performing the action, we captured the depth and color images of the effect. This procedure was repeated 10 times for each object, action, and tool. At the end of this scenario, we collected the 25600 color and depth images. This dataset can be employed on the following robot and computer vision problems: inferring human affordances, tool-action prediction for a coworking scenario, human-robot interactions, to mention a few.

\section{Data availablity}
A dedicated website~\footnote{\href{www.robotmultimodal.com}{www.robotmultimodal.com}}and a public repository~\footnote{\href{https://github.com/muratkirtay/icub-multimodal-dataset}{https://github.com/muratkirtay/icub-multimodal-dataset}} will be available to provide other researchers with access to datasets images, figures, and benchmarked results. Note that the content of this web site will frequently be updated to add new information regarding the datasets, benchmarked machine learning models, and host other researchers' results and models.

\subsubsection*{Acknowledgments}
We would like to thank Marcello Calisti, Mariangela Manti, Alessio Fasano,
Elisa Massi, Debora Zrinscak, Matteo Priorelli and Taimoor Shah Hassan for their invaluable help during various stages of the data acquisition procedure.

Guido Schillaci has received funding from the European Union’s Horizon 2020 research and innovation programme under the Marie Sklodowska-Curie grant agreement No. 838861 (Predictive Robots).

\bibliographystyle{authordate1}
\bibliography{icubMultisensorDataset}

\end{document}